\documentclass{article}

\usepackage[preprint]{neurips_2024}

\usepackage[utf8]{inputenc} 
\usepackage[T1]{fontenc}    
\usepackage{hyperref}       
\usepackage{url}            
\usepackage{booktabs}       
\usepackage{amsfonts}       
\usepackage{amsmath}
\usepackage{mathtools}
\usepackage{nicefrac}       
\usepackage{microtype}      
\usepackage{xcolor}         
\usepackage{multirow}
\usepackage{graphicx}
\usepackage{algorithm}
\usepackage{algorithmic}
\usepackage{xcolor}
\usepackage[table,xcdraw]{colortbl}

\def\bzero{{\mathbf 0}}

\def\bepsilon{{\mathbf \epsilon}}
\def\btheta{{\mathbf \theta}}

\def\bc{{\mathbf c}}
\def\be{{\mathbf e}}
\def\bh{{\mathbf h}}

\def\bw{{\mathbf w}}

\def\bz{{\mathbf z}}

\def\bA{\mathbf A}

\def\bI{{\mathbf I}}

\def\bW{{\mathbf W}}
\def\bX{{\mathbf X}}
\def\bY{{\mathbf Y}}

\def\sR{{\mathbb R}}

\def\sE{{\mathbb E}}

\def\gN{{\mathcal{N}}}

\def\gV{{\mathcal{V}}}

\def\gE{{\mathcal{E}}}
\def\gT{{\mathcal{T}}}
\def\gL{{\mathcal{L}}}
\def\gG{{\mathcal{G}}}

\usepackage[normalem]{ulem}
\newcounter{ToDo}
\newcounter{gaocomm} 
\newcounter{Note}
\definecolor{blue-violet}{rgb}{0.00,0.75,0.90}
\definecolor{mygreen}{rgb}{0.0, 0.5, 0.0}
\definecolor{awesome}{rgb}{1.0, 0.13, 0.32}
\definecolor{bostonuniversityred}{rgb}{1.0, 0.0, 0.0}

\title{Unleash Graph Neural Networks from Heavy Tuning}

\author{%
  Lequan Lin\thanks{Equal contribution. Lequan Lin is the corresponding author.} \\
  Discipline of Business Analytics, the University Business School, University of Sydney\\
  \texttt{lequan.lin@sydney.edu.au} \\
  \And
  Dai Shi$^*$ \\
  Discipline of Business Analytics, the University Business School, University of Sydney\\
  \texttt{dai.shi@sydney.edu.au} \\
  \AND
  Andi Han \\
  Center for Advanced Intelligence Project (AIP), Riken.\\
  \texttt{andi.han@riken.jp} \\
  \And
  Zhiyong Wang \\
  School of Computer Science, the University of Sydney\\
  \texttt{zhiyong.wang@sydney.edu.au} \\
  \And
  Junbin Gao \\
  Discipline of Business Analytics, the University Business School, University of Sydney\\
  \texttt{junbin.gao@sydney.edu.au} \\
}

\begin{document}

\maketitle

\begin{abstract}
  Graph Neural Networks (GNNs) are deep-learning architectures designed for graph-type data, where understanding relationships among individual observations is crucial. However, achieving promising GNN performance, especially on unseen data, requires comprehensive hyperparameter tuning and meticulous training. Unfortunately, these processes come with high computational costs and significant human effort. Additionally, conventional searching algorithms such as grid search may result in overfitting on validation data, diminishing generalization accuracy. To tackle these challenges, we propose a graph conditional latent diffusion framework (GNN-Diff) to generate high-performing GNNs directly by learning from checkpoints saved during a light-tuning coarse search. Our method: (1) unleashes GNN training from heavy tuning and complex search space design; (2) produces GNN parameters that outperform those obtained through comprehensive grid search; and (3) establishes higher-quality generation for GNNs compared to diffusion frameworks designed for general neural networks.
\end{abstract}

\section{Introduction} \label{sec: intro}
Graph Neural Networks (GNNs) are deep-learning architectures tailored for analyzing graph-structured data, where capturing the relationships among observations is essential \cite{wu2020comprehensive,zhou2020graph}. Akin to the training of most deep learning architectures, GNN training requires iterative optimization algorithms such as stochastic gradient descent (SGD) and hyperparameter tuning. Except for the model-related hyperparameters such as the teleport probability in APPNP \cite{gasteiger2019predict}, the optimizer alone requires heavy tuning on learning rate, weight decay, etc. Besides, we notice that the choice of hyperparameters indeed has a nontrivial impact on the final prediction outcome, especially for unseen data.  While automated search algorithms like grid search are commonly used to reduce manual effort, they still necessitate careful design of the search space. A limited search space cannot guarantee promising configurations, whereas an overly broad search space incurs high computational and time costs.  This motivates us to develop an alternative method to replace heavy tuning while preserving the model accuracy.


Previous works on GNN training generally focus on two topics: GNN pre-training and GNN architecture search. GNN pre-training involves using node, edge, or graph level tasks to find appropriate initialization of GNN parameters before fine-tuning with ground truth labels \cite{hu2019strategies,hu2020gpt,lu2021learning}. This approach still inevitably relies on hyperparameter tuning to boost model performance. GNN architecture search aims to find appropriate GNN architectures given data and tasks \cite{gao2019graphnas,you2020design}. However, it was explored in \cite{shchur2018pitfalls} that under comprehensive hyperparameter tuning and proper training process, simple GNN architectures such as GCN may even outperform sophisticated ones. Unfortunately, this also comes with high computational costs and significant human effort. 

\textbf{Novelty and Contributions} To alleviate the burden of heavy hyperparameter tuning, we propose a graph conditional latent diffusion framework (GNN-Diff) to directly generate high-performing GNN parameters by learning from checkpoints saved during a coarse search within a much smaller search space. While parameter and network generation have been extensively studied in previous research \cite{erkocc2023hyperdiffusion,peebles2022learning,schurholt2022hyper,soro2024diffusion,wang2024neural}, the intrinsic relationship between data characteristics and network parameters remains underexplored. Our approach uses a task-oriented graph autoencoder to integrate data and graph structural information as a condition for GNN parameter generation. We validate through empirical experiments on node classification that GNN-Diff: (1) reduces the need of extensive tuning and complex search space design by providing an alternative method to sample reliable parameters directly; (2) produces GNN parameters that outperform those obtained through comprehensive grid search by fully exploring the underlying population of "good parameters"; and (3) incorporates graph guidance to establish superior GNN parameter generation compared to diffusion frameworks designed for general neural networks. The implementation code of GNN-Diff will be released after the review process.

\section{Preliminary}
\subsection{Graph Neural Networks}
 We denote an undirected graph with $N$ nodes as $\mathcal{G}\{\mathcal{V},\mathcal{E},\bA\}$, where $\gV$ and $\gE$ are sets of nodes and edges and $\bA \in \sR^{N \times N}$ is the adjacency matrix containing information of relationships. $\bA$ 
 can be either weighted or unweighted, normalized or unnormalized. The graph signals are stored in a matrix $\bX \in \sR^{N \times D_f}$, where $D_f$ is the number of features. Traditional Multi-Layer Perceptrons (MLPs) integrate data information solely from the feature direction for representation learning, overlooking potential connections between nodes. This limitation is addressed by Graph Neural Networks (GNNs), a deep-learning architecture that incorporates additional relationship information through graph convolutions. Typically, there are two types of GNNs: spatial GNNs that aggregate neighboring nodes with message passing \cite{hamilton2017inductive,kipf2017semisupervised,gasteiger2019predict} and spectral GNNs developed from the graph spectral theory \cite{defferrard2016convolutional,lin2023magnetic, zou2023simple}. With both types, the most important component of their architecture is the convolutional layer, which usually consists of graph convolution, linear transformation, and non-linear activation functions. For example, the very classic convolutional layer in GCN \cite{kipf2017semisupervised} is formulated as $\sigma(\bA \bX \bW)$, where graph convolution involves node aggregation with graph adjacency $\bA$, linear transformation is conducted with linear operator $\bW$, and $\sigma$ is the activation function. The GNN training process often involves learning parameters in the linear transformation and for some architectures such as ChebNet\cite{defferrard2016convolutional} and the graph convolution.




\subsection{Latent Diffusion Models}
Diffusion models generate a step-by-step denoising process that recovers data in the target distribution from random white noises. A typical diffusion model adopts a pair of forward-backward Markov chains, where the forward chain perturbs the observed samples eventually to white noises, and then the backward chain learns how to remove the noises and recover the original data. Assume that $\bw_0 \sim q_{\bw}(\bw_0)$ is the original data (e.g., vectorized network parameters) from the target distribution $q_{\bw}(\bw_0)$. Then, the forward-backward chains are formulated as follows.

\noindent\textbf{Forward Chain} For diffusion steps $t = 0, 1, ..., T$, Gaussian noises $\bepsilon \sim \gN(\textbf{0},\bI)$ are injected to $\bw_t$ until $q_{\bw}(\bw_T) \coloneqq \int q_{\bw}(\bw_T|\bw_0) q_{\bw}(\bw_0) d \bw_0 \approx \gN(\bw_T; \bzero, \bI)$. By the Markov property, one may jump to any diffusion steps via $\bw_t = \sqrt{\widetilde{\alpha}_t} \bw_0 + \sqrt{1-\widetilde{\alpha}_t} \bepsilon$, where $\widetilde{\alpha}_t = \prod_{i=1}^t (1 - \beta_i)$ with $\beta = \{\beta_1, \beta_2, ..., \beta_T\}$ is a pre-defined noise schedule.

\noindent\textbf{Backward Chain} The backward chain removes noises from $\bw_T$ gradually via a backward transition kernel $q_{\bw}(\bw_{t-1}|\bw_t)$, which is usually approximated by a neural network $p_\btheta(\bw_{t-1}|\bw_t)$ with learnable parameters $\btheta$. Here we use the framework of denoising diffusion probabilistic models (DDPMs). Thus, we alternatively find a denoising network $\epsilon_\btheta$ to predict noise injected at each diffusion step with the loss function 
\begin{equation}
    \gL_{\text{DDPM}} = \sE_{t,\bw_0\sim q_{\bw}(\bw_0),\bepsilon \sim \gN(\bzero,\bI)} \left\| \bepsilon - \bepsilon_\btheta\left(\sqrt{\widetilde{\alpha}_t} \bw_0 + \sqrt{1-\widetilde{\alpha}_t} \bepsilon, t\right) \right\|^2.
\end{equation}

\noindent\textbf{Latent Diffusion} Classic two-layer GNNs such as GCN and ChebNet usually contain 10k to 500k parameters for benchmark graph datasets with 1k to 4k features. So generating GNN parameters directly with diffusion models may lead to slow training and long inference time. As a common solution to such problems, latent diffusion models (LDM) \cite{rombach2022high, soro2024diffusion, wang2024neural} first learn a parameter autoencoder (PAE) to convert the target parameters $\bw_0 \in \sR^{D_w}$ to a low-dimensional latent space $\bz_0 = \text{P-Eecoder}(\bw_0) \in \sR^{D_p}$ with $D_p \ll D_w$, then generate in the latent space before reconstructing the original target with a sufficiently precise decoder. The loss function is usually formulated as the mean squared error (MSE) between original data $\bw_0$ and reconstructed data $\text{P-Decoder}(\bz_0)$.  

\section{Related Works}

\textbf{Hypernetworks} Network generation usually refers to the generation of neural network parameters. Hypernetwork \cite{ha2016hypernetworks,stanley2009hypercube} is an early concept of deep learning technique for network generation and now serves as a general framework that encompasses many existing methods. A hypernetwork is a neural network that learns how to predict the parameters of another neural network (target network). The input of a hypernetwork may be as simple as parameter positioning embeddings \cite{stanley2009hypercube}, or more informative, such as encodings of data, labels, and tasks \cite{deutsch2018generating,ha2016hypernetworks,ratzlaff2019hypergan, zhang2018graph,zhmoginov2022hypertransformer}. 

\noindent\textbf{Generative Modelling for Network Generation} With the emergence of generative modelling, studies on distributions of network parameters have gained their popularity in relevant research. A pioneering work of \cite{unterthiner2020predicting} showed that simply knowing the statistics of parameter distributions enabled us to predict model accuracy without accessing the data. Enlightened by this finding, two following works \cite{schurholt2022hyper,schurholt2021self} studied the reconstruction and sampling of network parameters from their population distribution. The similar idea was also investigated by \cite{peebles2022learning}, who proposed that network parameters corresponding to a specific metric value of some specific data and tasks could be generated with a transformer-diffusion model trained on tremendous amount of model checkpoints collected from past training on a group of data and tasks. This method, however, relies highly on the amount and quality of model checkpoints fed to the generative model. A more network- and data-specific approach, p-diff \cite{wang2024neural}, alternatively collects checkpoints from the training process of the target network and uses an unconditional latent diffusion model to generate better-performing parameters. Another recent work, D2NWG \cite{soro2024diffusion}, adopts dataset-conditioned latent diffusion to generate parameters for target networks on unseen datasets. Diffusion-based network generation has also been widely applied to other applications, such as meta learning \cite{nava2022meta, zhang2024metadiff} and producing implicit neural-filed for 3D and 4D synthesis \cite{erkocc2023hyperdiffusion}.

\noindent\textbf{Parameter-free Optimization, Bayesian Optimization and Meta Learning.}
There emerge many recent works on parameter-free optimization, such as \cite{defazio2023learning,ivgi2023dog,mishchenko2023prodigy} that automatically set step-size based on problem characteristics. However, most if not all methods require non-trivial adaptation of existing optimizers and show theoretical guarantees only for convex functions. Bayesian optimization \cite{snoek2012practical} and meta learning \cite{franceschi2018bilevel} on the other hand, are more advanced approaches for tuning hyperparameters in a more informed way than grid and random search. 



\section{Graph Neural Network Diffusion (GNN-Diff)}
\begin{figure}[t!]
    \centering
    \scalebox{1}{
    \includegraphics[width=1\linewidth]{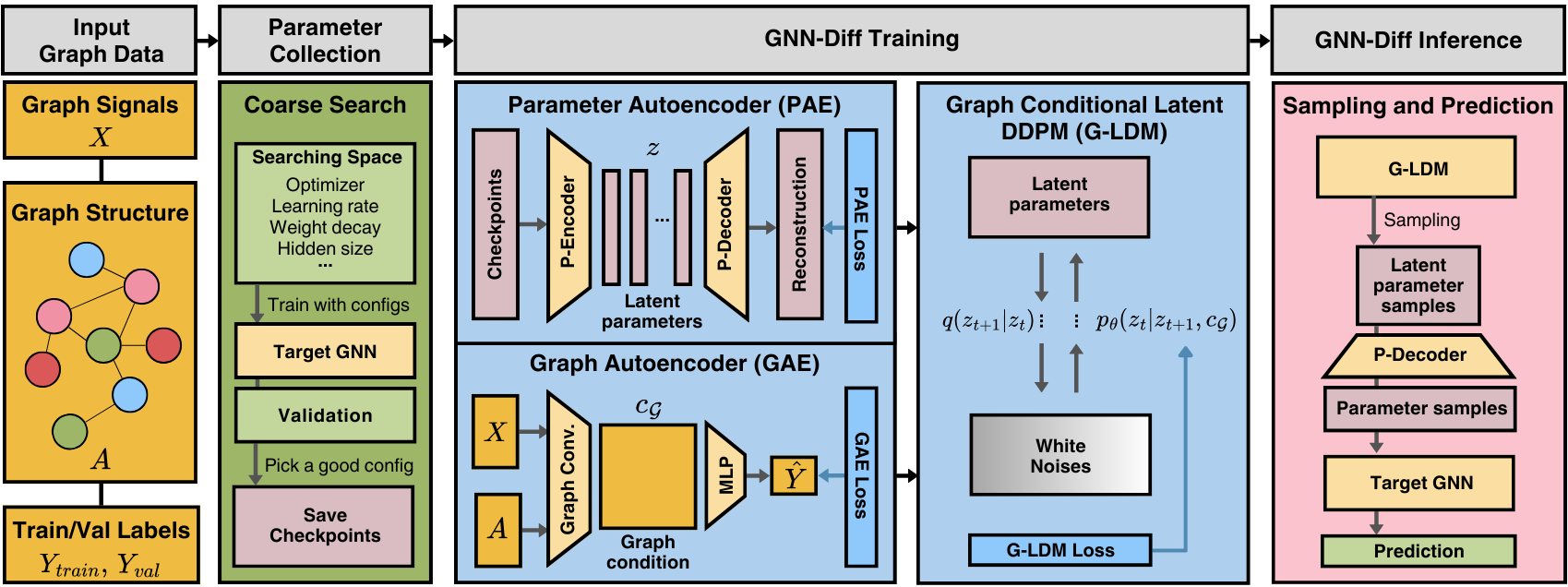}}
    \caption{GNN-Diff Overview. (1) Input graph data: input graph signals, adjacency matrix, and train/validation ground truth labels. (2) Parameter collection: a coarse search with a small search space is conducted to select an appropriate hyperparameter configuration, which is then used to collect model checkpoints. (3) Training: PAE and GAE are firstly trained to produce latent parameters and graph conditions, and then G-LDM learns how to recover latent parameters from white noise with graph conditions. (4) Inference: after sampling latent parameters from G-LDM, GNN parameters are reconstructed with a PAE decoder and returned to target GNN for prediction.}
    \label{fig:method}
\end{figure}

GNN-Diff takes 4 steps before final prediction on unseen data: (1) inputting graph data, (2) collecting parameters with coarse search, (3) training, and (4) sampling and reconstructing parameters. Training involves the training of three modules: parameter autoencoder (PAE), graph autoencoder (GAE), and graph conditional latent diffusion (G-LDM). We provide a visualized overview of GNN-Diff in Figure~\ref{fig:method}. Since input data and PAE have been previously discussed with preliminaries, we predominantly focus on the remaining components in this section. The pseudo codes of training and inference algorithms are provided in Appendix \ref{apx: pseudo}.

\subsection{Parameter Collection with Coarse Search}
To produce high-performing GNN parameters, we need a set of high-quality parameter samples such that the diffusion module can generate from their underlying population distribution. It is quite intuitive that good model performance is associated with appropriate selection of hyperparameters. So, we start with a coarse search with a relevantly small search space to determine a suitable configuration (details on the search space are discussed in Section \ref{sect:experiments}). This includes the selection of hyperparameters in the optimizer such as learning rate, model architecture such as hidden size, and other model-specific factors such as the teleport probability $\alpha$ in APPNP \cite{gasteiger2019predict}. Since parameter initialization may also influence the training outcome, we provide 10 random initializations for training with each configuration. A good configuration from coarse search is defined as the one leading to the best validation accuracy. After the coarse search, we save model checkpoints from the training process with the selected configuration as parameter samples. We discard the parameters from some start-up epochs where convergence has not been achieved. Parameters are reorganized and vectorized before they are further processed by PAE. 

\subsection{Graph Autoencoder (GAE)}\label{gae}
GAE is designed to encode graph information from graph signals $\bX$ and structure $\bA$, which will be employed as a condition for GNN generation. In this paper, we mainly focus on the node classification task to evaluate our method, so the GAE structure is oriented by the specific task, which means GAE encoder aims to produce a graph representation that works well on node classification. Straightforwardly, a GNN architecture should be considered. The general principle of GNN propagation is to update the node feature by aggregating its neighboring information. Thus, GNN graph representation has similar features for connected nodes, which will eventually lead to similar label prediction. This is well-suited for homophilic graphs, where connected nodes are normally from the same classes. However, for heterophilic graphs, where nodes with different labels are prone to be linked, such representation leads to an even worse classification outcome than MLP \cite{han2024from,zheng2022graph}. Hence, enlightened by some previous works \cite{shao2023unifying,thorpe2022grand,zhu2020beyond}, we introduce the following GAE encoder, which is capable of handling both homophilic and heterophilic graphs: 
\begin{equation}\label{eq_gae_encoder}
    \bh = \text{G-Encoder}(\bX,\bA) = \mathrm{Concat}\left(\bA^2\bX\bW_1,\bA\bX\bW_2, \bX\bW_3 \right) \bW_4  \in \sR^{N \times D_p},
\end{equation}
which is composed of the concatenation of a two-layer GCN, a one-layer GCN, and an MLP, followed by a linear transformation $\bW_4 \in \sR^{D_\mathcal{G} \times D_p}$, where $D_\mathcal{G}$ and $D_p$ are the dimensions of the concatenation and the latent parameter representation, respectively. $\text{G-Decoder}(\bh) = \bh \bW_5 \in \sR^{D_c}$ is a single layer MLP with $D_c$ being the number of node classes. Thus graph information is fully encapsulated in the output of the encoder. Finally, GAE is trained with cross entropy loss on node labels. The graph condition is then given by mean-pooling the node features
\begin{equation}
    \bc_{\mathcal{G}} = \frac{1}{N} \sum_{i = 1}^N \bh(i) \in \sR^{D_p},
\end{equation}
where $\bh(i)$ is the i-th row of graph representation. The purpose of applying mean pooling on nodes is to construct an overall graph condition regardless of the graph size, which will later be combined with latent representation $\bz_0 \in \sR^{D_p}$ as input of diffusion denoising network. We highlight that our method can be naturally extended to other graph tasks by altering the design and loss function of GAE.

\subsection{Graph Conditional Latent DDPM (G-LDM)}
G-LDM is developed from LDM with additional graph conditions from GAE to provide consolidated guidance on GNN generation. The generative target of G-LDM is latent GNN parameters $\bz_0$. The forward pass of G-LDM is the same as traditional DDPM, while the backward pass uses a graph conditional backward kernel for data recovering from white noises. The corresponding G-LDM loss function is given by
\begin{equation}
    \gL_{\text{G-LDM}} = \sE_{t,\bz_0\sim q_{\bz}(\bz_0),\bepsilon \sim \gN(\bzero,\bI)} \left\| \bepsilon - \bepsilon_\btheta\left(\sqrt{\widetilde{\alpha}_t} \bz_0 + \sqrt{1-\widetilde{\alpha}_t} \bepsilon, t, \bc_{\gG}\right) \right\|^2.
\end{equation}
For both PAE and denoising network $\bepsilon_\btheta$, we adopt the \texttt{Conv1d}-based architecture in \cite{wang2024neural}. We include the graph condition by using the sum of $\bz_t$, $\bc_\mathcal{G}$, and $\text{Embedding}(t)$ as input to $\bepsilon_\btheta$. Our method is flexible with PAE and $\bepsilon_\btheta$ architectures. So, we choose an architecture that has been empirically validated by previous research. We observe in experiments that the most important factor in high-quality generation is to reduce parameter dimension sufficiently while ensuring reconstruction tightness. This can be done by setting the appropriate kernel size and stride for \texttt{Conv1d} layers in PAE. We discuss more details about PAE and $\bepsilon_\btheta$ architecture and how to ensure high-quality generation in Appendix \ref{apx: conv_arch} and Appendix \ref{apx: pae_hyper}, respectively.

\subsection{Sampling, Reconstruction, and Prediction} 
The last step before the final prediction is to obtain parameters and reconstruct the target GNN. G-LDM generates samples of latent parameters from white noises with the learned denoising network. Next, the learned PAE decoder is applied to reconstruct parameters back to the original dimension. These parameters are returned to the target GNN thus we have a group of GNNs with generated parameters. Lastly, we determine the optimal GNN depending on validation performance, which will be used for future predictions.

\section{Experiments and Discussions} \label{sect:experiments}
\subsection{Experimental Setup}
\noindent \textbf{Dataset and Target GNNs} We evaluate our method with 4 benchmark graph datasets, Cora \cite{yang2016revisiting}, Citeseer \cite{yang2016revisiting}, Actor \cite{Pei2020Geom-GCN}, and Chameleon \cite{rozemberczki2021multi}. Cora and Citeseer are homophilic graphs, while Actor and Chameleon are heterophilic graphs. Data split follows the original works. We use only the first train/val/test mask for Actor and Chameleon. 7 most representative GNNs are selected as target models, including GCN\cite{kipf2017semisupervised}, APPNP \cite{gasteiger2019predict}, SAGE \cite{hamilton2017inductive}, ChebNet \cite{defferrard2016convolutional}, GIN \cite{xu2019how}, GAT \cite{velivckovic2017graph}, and H2GCN \cite{zhu2020beyond}. In particular, we evaluate GCN with a single layer (GCN1) and two layers (GCN2) to explore the ability of GNN-Diff to generate parameters for models with different depths. H2GCN is H2GCN1 from its original paper. All other architectures are two-layer GNNs. We include more details of datasets in Appendix \ref{apx: data} and target GNNs and why they are selected in Appendix \ref{apx:GNN}.

\noindent \textbf{Experiment Details} All experiments are run on a single NVIDIA 4090 GPU with 64GB memory. For each target GNN, we conduct coarse search with a search space of 20-80 hyperparameter configurations, whereas the grid search takes a more comprehensive search space of 144-1269 configurations. A list of search spaces can be found in Appendix \ref{apx: search space}. For each configuration, we train the target GNN with 200 epochs to ensure convergence and 10 parameter initializations to minimize the influence of randomness. GNN-Diff training is designed as a training flow on 3 respective modules: GAE, PAE, and G-LDM. To alleviate hyperparameter tuning, we designed GNN-Diff modules to require minimal tuning. In fact, the only hyperparameters to consider are the PAE kernel size, fold rate, and the number of training epochs. These are determined by the target GNN parameter size with fairly simple rules. We provide relevant training details in Appendix \ref{apx: pae_hyper}.  
For evaluation on the test set, we generate 100 latent parameter representations with G-LDM, which are then reconstructed by the learned $\text{P-Decoder}$ and recovered to the target GNN. The generated model with the best validation accuracy will be tested. 


\noindent \textbf{Reproducibility and Comparability} Random seeds of all packages in the implementation code are set as 42 to mitigate randomness. The same hyperparameter setting and training process are applied to generative models for fair comparison in Subsection \ref{exp_graph_cond}.  

\subsection{Results on Node Classification}\label{sec: result_nc}

\begin{table}[h!]
\centering
\caption{Test accuracy of GNNs selected by comprehensive grid search (G.S.), coarse search (C.S.), and generated by GNN-Diff (G-D). All models are selected based on validation accuracy. Best results are highlighted in \textbf{bold}.}
\label{table:node_result}
\renewcommand{\arraystretch}{1.1}
\setlength{\tabcolsep}{11pt}
\scalebox{0.67}{
\begin{tabular}{c|ccc|ccc|ccc|ccc}
\toprule
\multirow{2}{*}{Model} & \multicolumn{3}{c|}{Cora}      & \multicolumn{3}{c|}{Citeseer}           & \multicolumn{3}{c|}{Actor}     & \multicolumn{3}{c}{Chameleon}                    \\ \cline{2-13} 
                       & G.S.  & C.S   & G-D       & G.S.           & C.S   & G-D       & G.S.  & C.S   & G-D       & G.S.           & C.S            & G-D       \\ \hline
GCN1                   & 78.70 & 78.10 & \textbf{79.00} & 70.30          & 69.50 & \textbf{70.60} & 29.67 & 29.74 & \textbf{29.80} & \textbf{57.24} & 55.26          & \textbf{57.24} \\ \hline
GCN2                   & 80.90 & 81.80 & \textbf{82.10} & 72.50          & 70.20 & \textbf{72.70} & 31.38 & 30.13 & \textbf{31.84} & 65.57          & 65.35          & \textbf{66.01} \\ \hline
APPNP                  & 82.20 & 82.70 & \textbf{83.00} & 72.40          & 72.00 & \textbf{73.40} & 35.20 & 35.00 & \textbf{36.32} & 58.77          & 58.11          & \textbf{60.09} \\ \hline
SAGE                   & 80.70 & 80.90 & \textbf{81.90} & 70.90          & 67.10 & \textbf{71.20} & 36.71 & 35.66 & \textbf{37.63} & \textbf{61.99} & 60.09          & 61.84          \\ \hline
ChebNet                & 79.50 & 79.90 & \textbf{80.80} & 70.60 & 68.80 & \textbf{70.69}          & 36.58 & 36.12 & \textbf{36.90} & 58.99          & 57.46          & \textbf{59.86} \\ \hline
GIN                    & 75.80 & 76.30 & \textbf{76.70} & \textbf{68.40} & 68.10 & 63.90          & 26.25 & 25.92 & \textbf{26.97} & \textbf{38.82} & 33.99          & 33.33          \\ \hline
GAT                    & 82.20 & 81.40 & \textbf{83.50} & \textbf{73.00} & 71.50 & 71.00          & 30.53 & 29.80 & \textbf{30.72} & 63.82          & \textbf{64.47} & \textbf{64.47} \\ \hline
H2GCN                  & 80.90 & 80.20 & \textbf{81.40} & 70.40          & 71.20 & \textbf{72.00} & 33.49 & 33.49 & \textbf{35.26} & 53.29          & 51.75          & \textbf{53.95} \\ \bottomrule
\end{tabular}}
\end{table}

In Table \ref{table:node_result}, we present the results of GNN-Diff on generating GNNs for node classification. We compare the test accuracy of three models selected respectively from grid search, coarse search, and GNN-Diff generation based on validation accuracy. For simplicity, we will refer them as grid model, coarse model, and GNN-Diff model. Overall, GNN-Diff models outperform grid and coarse models on most GNN architectures and datasets. On Cora and Actor, GNN-Diff models successfully achieve the best test accuracy for all target GNNs. GNN-Diff models improve the test accuracy from 0.20\% to 1.30\% on Cora and from 0.06\% to 1.77\% on Actor. On Citeseer and Chameleon, GNN-Diff models also produce good test results for most GNN architectures. Notably, GNN-Diff boosts the test performance for H2GCN on Citeseer by 1.60\% and for APPNP on Chameleon by 1.32\%. However, GNN-Diff fails to outperform grid results for GIN and GAT on Citeseer and for GIN and SAGE on Chameleon. We have tracked the validation and test accuracy of GIN during the training for grid search, and noticed that good accuracy in grid search could be considered as outliers in overall GIN performance. They appeared abruptly in the training history and had much higher value than neighboring epochs. Thus, it is hard to capture the corresponding parameters based on coarse search samples. We suppose it is reasonable to treat this as special situation. For GAT on Citeseer and SAGE on Chameleon, heavy parameterization of their architectures lays the burden on PAE reconstruction. We suggest that a more sophisticated architecture of PAE may be helpful, though the current architecture still works well on these two GNNs for other datasets.

The grid search accuracy we report here for some GNNs may be lower than other relevant works. This is majorly caused by different experiment setting. The grid model evaluated by test set is the model with highest validation accuracy among all others from the entire comprehensive grid search process. Some relevant works run GNNs with each individual hyperparameter configuration, and report the best test accuracy among configurations. Validation data is only used for model selection during the training process with each configuration.

\subsection{Visualizations}
\begin{figure}[h!]
    \centering
    \scalebox{1}{
    \includegraphics[width=1\linewidth]{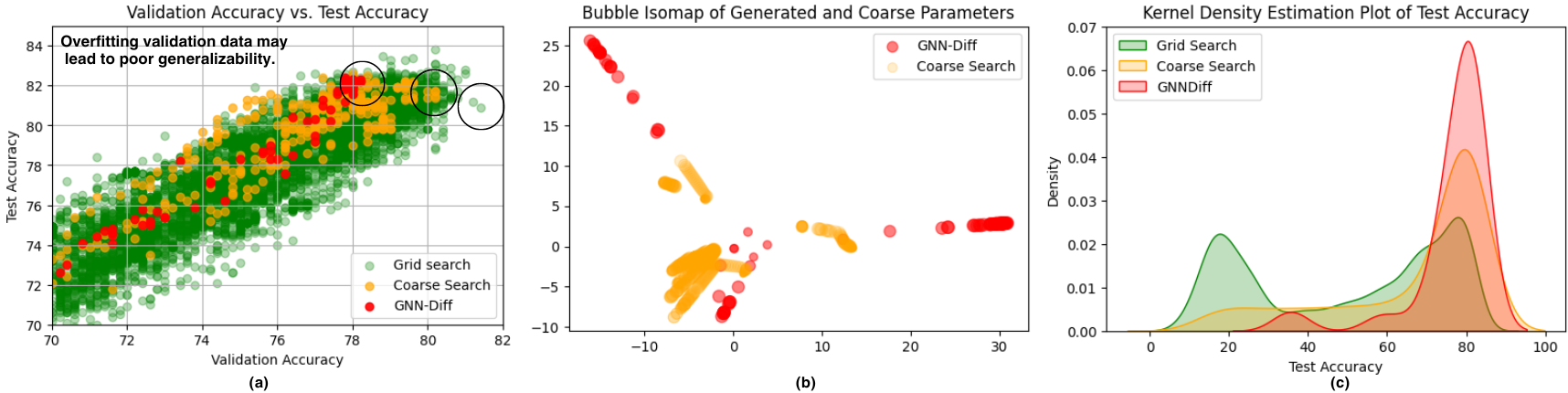}}
    \caption{(a) Scatter plot between validation and test accuracy. Black circles indicate the position of the final models selected by three methods. (b) Visualization of generated and coarse parameters after dimension reduction with isomap. Each bubble represents one parameter sample, and the bubble size shows the corresponding test accuracy. (c) The kernel density estimation plot shows the distribution of test accuracy associated with parameters learned with grid search and coarse search and generated with GNN-Diff. }
    \label{fig:visual}
\end{figure}

We further investigate the experiment results by visualization. With coarse search, we only visualize the saved checkpoints for GNN-Diff training to be directly compared with models generated by GNN-Diff. We use GCN2 on Cora as an example and try to answer three questions:

\noindent \textbf{Q1. Why are some coarse search outcomes better than grid search in Table \ref{table:node_result}?}

\noindent \textbf{A1. Comprehensive grid search may overfit validation set.} 

In Figure \ref{fig:visual} (a), we present a scatter plot between validation and test accuracy for partial grid (green), coarse (orange), and GNN-Diff (red) models that achieve at least 70\% prediction accuracy on both validation and test set. In our experiments, the models from three methods with the highest validation accuracy are selected for evaluation on test data. We use three black circles to indicate the position of the final models. Clearly, the grid model with the highest validation accuracy has the lowest test accuracy compared to the other two methods. This is a sign of overfitting on validation data, which eventually results in poor generalizability. 

Although grid search generally guarantees to find a promising hyperparameter configuration, its reliance on validation may be counterproductive sometimes. Having a smaller search space or using other search algorithms, such as random search, may solve the problem. But how do we know the appropriate search space size? In addition, nearly 80\% of our experiment results show that grid search can find better generalized model than coarse search, hence reducing search space or random search may not be good options in most cases. This shows the superiority of our method, which only requires samples from a good configuration and then further optimizes the model performance by exploring the underlying population distribution. This avoids finding a configuration that is optimal for validation but not for unseen data.

\noindent \textbf{Q2. What is the relationship between coarse models and GNN-Diff models?} 

\noindent \textbf{A2. GNN-Diff parameters follow the coarse distribution and explore into the population.}

As we know, the parameters of coarse models are the training data of GNN-Diff. So, the parameters generated by GNN-Diff are expected to follow a similar distribution as coarse model parameters. In Figure \ref{fig:visual} (b), we use a bubble isometric mapping plot to visualize coarse (orange) and GNN-Diff (red) parameter distributions via dimensionality reduction. Isometric mapping (Isomap) seeks to preserve the geodesic distances between all pairs of data points in the 2D embedding. It is particularly helpful for visualizing high-dimensional parameters by projecting them onto a low-dimensional space while maintaining the underlying geometry. Additionally, we plot each model point as bubbles with bubble size representing the corresponding test accuracy. 

We observe that GNN-Diff parameters follow the pattern of the coarse distribution (a firework-like expanding spread with three directions). Besides, the generation further explores the potential population distribution by extending in two of the three directions. Notably, the generated models close to the center of coarse distribution are associated with lower accuracy, whereas the extended ones have better prediction outcomes. This may validate our conjecture that GNN-Diff has the ability to approximate population parameter distribution with a good hyperparameter configuration and generate better-performing samples from it.

\noindent \textbf{Q3. How can GNN-Diff models produce better prediction result than coarse and grid models?}

\noindent \textbf{A3. GNN-Diff leads to more centered accuracy distribution around a high accuracy region.}

In Figure \ref{fig:visual} (c), we provide the Kernel Density Estimation (KDE) plot of test accuracy distributions for grid models (green), coarse models (orange), and GNN-Diff models (red). The test accuracy of grid models shows a broader distribution with two peaks, one around 20\% and the other one around 80\%. Grid search, while thorough, spends much unnecessary time and computation on training with sub-optimal or even unsatisfactory configurations. Coarse models present a test accuracy distribution with a peak around 80\% but with some variability. Two factors have contributed to this distribution. Firstly, coarse models are collected from a good hyperparameter configuration of coarse search. Secondly, we discard some models from start-up epochs that are generally associated with low test accuracy. The test accuracy distribution of GNN-Diff models has a sharp peak at a comparably higher level, implying high-performing outcomes with less variability than the other two methods. This means GNN-Diff consistently finds models with comparably better generalizability, outperforming both grid search and coarse search in terms of finding high-quality parameters.

\subsection{Is Graph Condition Useful for GNN Generation?}\label{exp_graph_cond}
Here we focus on whether incorporating graph condition is useful for GNN generation. Previous works have discussed using data and tasks as conditions, but only for the purpose of tailoring their methods for future unseen data or similar tasks \cite{nava2022meta,soro2024diffusion,zhang2024metadiff}. The role of intrinsic data characteristics in network generation is still unexploited.  Nevertheless, with graph neural networks, the graph structure is very essential in the network architecture and has a consequential impact on graph signal processing. So, only considering data (graph signals) as the generative condition is insufficient.  Previous study \cite{nava2022meta} includes predictive tasks as the condition by encoding their language descriptions, which requires additional effort to form appropriate descriptions and embedding. Alternatively, we propose a more straightforward way via specific task supervision on condition construction. Therefore, we design the GAE discussed in Subsection \ref{gae}, which incorporates both graph data and structural information in the generative condition and is trained with the task loss function. We conduct two experiments: (1) a comparison experiment between GNN-Diff and p-diff \cite{wang2024neural} to show how graph condition improves GNN generation quality; (2) an ablation study on GNN-Diff generative condition to validate the current GAE architecture on homophilic and heterophilic graphs. In the first experiment, we chose p-diff because it has the same PAE and $\bepsilon_\btheta$ as GNN-Diff and no generative condition. Hence, we do not need to consider different performances caused by different generative model architectures. Both methods are trained with the same hyperparameters and experiment settings. For both experiments, we conduct sampling and prediction 5 times and report the best results for all methods/conditions to offset the influence of randomness in diffusion generation and ensure fair comparison. 

\textbf{Comparison between GNN-Diff and p-diff}

\begin{figure}[h!]
    \centering
    \includegraphics[width=1\linewidth]{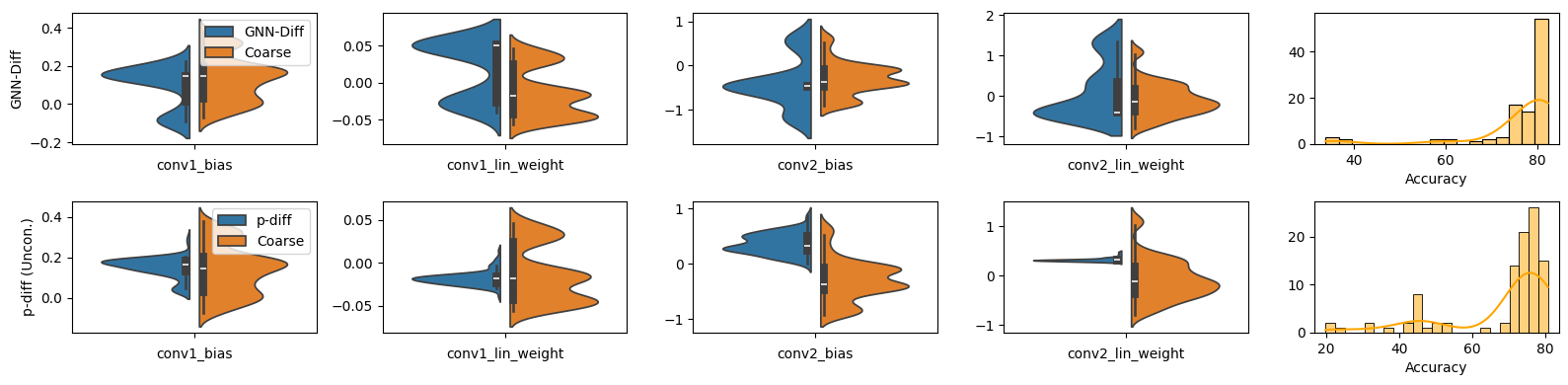}
    \caption{Distribution of parameters generated with/without graph condition for GCN2 on Cora. Last two plots on the right show the distribution of corresponding test accuracy.}
    \label{fig:compare_dist}
\end{figure}

In Figure \ref{fig:compare_dist}, we compare the generated parameter and test accuracy distributions with and without graph conditions. This experiment is conducted with GCN2 on Cora. Based on the implementation of \texttt{PyTorch Geometric}, GCN2 has 4 learnable components, including biases and weights in two graph convolutional layers. For visualization purposes, we pick one parameter from each component and plot its distributions with sample models from coarse search (orange), GNN-Diff generation (blue, top row), and p-diff generation (blue, bottom row). It is clear that GNN-Diff generation captures the sample distribution and extends the sample distribution range to explore potential population distribution. By contrast, p-diff generation produces much centered distribution but fails to cover the sample distribution. This indicates that graph condition provides a useful guidance on GNN generation, and helps the generative model to better learn the target distribution. We also provide histograms of test accuracy distribution for both methods. p-diff models obtain around 20\% to 80\% accuracy when being applied to test data. GNN-Diff models, on the other hand, have accuracy more centered around 80\%. Also, the worst-performing model generated by GNN-Diff still outperforms many p-diff models. Although high accuracy can be achieved by p-diff, the generation quality is unstable without the guidance of graph conditions.

\textbf{Ablation Study on Generative Condition}

\begin{figure}[h!]
    \centering
    \includegraphics[width=0.6\linewidth]{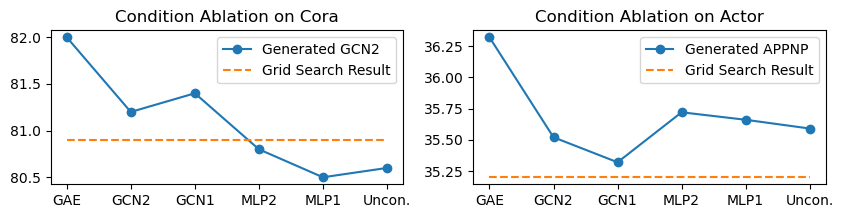}
    \caption{Ablation study on generative condition. GNN-Diff employs graph conditions provided by GAE. The model is equivalent with p-diff \cite{wang2024neural} when no condition is applied (Uncon.).}
    \label{fig:ablation}
\end{figure}

In Figure \ref{fig:ablation}, we present the test accuracy of GNN generation with different conditions. GAE refers to the architecture discussed in Subsection \ref{gae}, which is adopted by GNN-Diff. According to Equation (\ref{eq_gae_encoder}), GAE encoder combines GCN and MLP to handle both homophilic and heterophilic graphs. In this experiment, we apply ablation on graph conditions. GCN1 and GCN2 represent graph conditions with only graph convolution-based representation. MLP2 and MLP1 represent conditions with only data information. Lastly, we include no condition (p-diff) as the baseline.  Different target GNNs are considered to show the generalizability of the conclusion. On the homophilic graph, Cora, GAE leads to the best test result, while conditions without graph structural information (MLP2 and MLP1) and no condition fail to outperform grid search results. On heterophilic graphs, Actor, MLP2, MLP1, and no condition produced better results than graph convolution-based conditions (GCN2 and GCN1). However, in GAE, the combination of graph convolution and MLP still generates the best prediction outcome. Therefore, we conclude that the current GAE architecture consistently leads to high-performing parameters regardless of the graph type.

\subsection{Time Efficiency}
\begin{figure}[h!]
    \centering
    \includegraphics[width=1\linewidth]{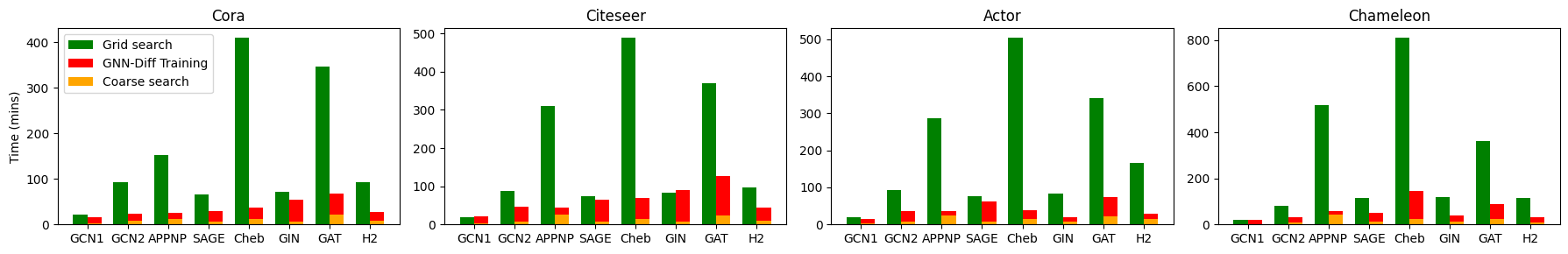}
    \caption{Time costs of comprehensive grid search, coarse search, and GNN-Diff training. The sampling time of GNN-Diff is negligible (less than 10 seconds to generate 100 GNNs).}
    \label{fig:time}
\end{figure}

Last but not least, we would like to discuss the time efficiency of GNN-Diff. In Figure \ref{fig:time}, we use a bar chart to show the time of grid search (green), coarse search (orange), and GNN-Diff training (red) for different target GNNs and datasets. We stack bars of coarse search and GNN-Diff training as the time cost of the entire GNN-Diff process. Inference time is not shown as it is negligible (usually less than 10 seconds). The time is estimated based on the average search iteration time and average epoch time for three GNN-Diff modules. Implementation time may vary due to factors such as data pre-processing and validation. According to the plots, it is very clear that the GNN-Diff method is more efficient than a comprehensive grid search. The advantage is less obvious for simple GNN architectures such as GCN1, for which only a small search space is needed for grid search. However, it is very valuable for architectures with large hyperparameter space such as GAT and GNNs with cumbersome graph convolution such as APPNP.

\section{Conclusion} \label{sec: con}
In this paper, we proposed a graph conditional latent diffusion framework, GNN-Diff, to generate high-performing GNN parameters by learning from checkpoints saved with a light-tuning coarse search. We validate with empirical experiments that our method is an efficient alternative to costly search algorithms and generates better prediction outcomes than a comprehensive grid search on unseen data. We have also shown that, by incorporating a carefully designed task-orientated graph condition in the generation process, our method establishes higher-quality generation for GNNs than diffusion methods designed for general network generation. We majorly focus on the node classification task to validate our method, though it can be naturally extended to other graph tasks. Future works may involve the implementation of other tasks and solutions to reconstructing heavy parameterized GNNs.

\clearpage 
\bibliographystyle{plain}
\bibliography{reference}

\clearpage
\appendix

\section{Pseudo Code of GNN-Diff} \label{apx: pseudo}





\begin{algorithm}[h!]
\caption{Algorithm of GNN-Diff Network Generation}\label{algo: overall}
\textbf{Input:} Graph condition $\bc_\gG$ from Algorithm \ref{algo: gae}; the learned P-Decoder$(\cdot)$ from Algorithm \ref{algo: pae}; the learned denoising network $\bepsilon_\btheta$ from Algorithm \ref{algo: gldm}.

\textbf{Output:} Generated parameters $\hat{\bw}_0$ that can be returned to target GNN for prediction.

\begin{algorithmic}[1]
\STATE Run Algorithm \ref{algo: sample} to sample latent parameter $\hat{\bz}_0$
\STATE Reconstruct parameters with the learned PAE decoder: $\hat{\bw}_0 = \text{P-Decoder}(\hat{\bz}_0)$

\end{algorithmic}
\end{algorithm}

\vspace{0.2cm}

\begin{algorithm}[h!]
\caption{Training Algorithm of GAE} \label{algo: gae}
\textbf{Input:} Graph signals $\bX$; graph structure $\bA$; graph label $\bY$; graph training mask $M_{train}$; number of training epochs $E_{\text{GAE}}$.

\textbf{Output:} Graph condition $\bc_\mathcal{G}$.
\begin{algorithmic}[1]

\FOR{$i = 1, 2, ..., E_{GAE}$}
\STATE Compute G-Encoder: $\bh = \mathrm{Concat}\left(\bA^2\bX\bW_1,\bA\bX\bW_2, \bX\bW_3 \right) \bW_4$
\STATE Compute G-Decoder: $\hat{\bY} = \mathrm{Softmax}(\bh \bW_5)$
\STATE Take gradient step and update $\bW = \{\bW_1, \bW_2, \bW_3, \bW_4, \bW_5\}$:
\begin{equation*}
    \nabla_{\bW} \mathrm{CrossEntropy}(\bY \odot M_{train},\hat{\bY} \odot M_{train})
\end{equation*}
\ENDFOR

\STATE Compute graph representation $\bh = \mathrm{Concat}\left(\bA^2\bX\bW_1,\bA\bX\bW_2, \bX\bW_3 \right) \bW_4$
\STATE Conduct mean pooling to obtain graph condition $\bc_{\mathcal{G}} = \frac{1}{N} \sum_{i = 1}^N \bh(i)$.

\end{algorithmic}
\end{algorithm}

\vspace{0.2cm}

\begin{algorithm}[h!]
\caption{Training Algorithm of PAE} \label{algo: pae}
\textbf{Input:} Vectorized parameters collected from coarse search $\bw_0 \sim q_{\bw}(\bw_0)$; number of training epochs $E_{\text{PAE}}$.

\textbf{Output:} Latent parameter representation $\bz_0 \sim q_{\bz}(\bz_0)$; the learned decoder $\text{P-Decoder}(\cdot)$.

\begin{algorithmic}[1]
\FOR{$i = 1, 2, ..., E_{\text{PAE}}$}
\STATE Compute latent parameter representation $\widetilde{\bz}_0 = \text{P-Encoder}(\bw_0)$
\STATE Compute PAE decoder for reconstruction $\widetilde{\bw}_0 = \text{P-Decoder}(\widetilde{\bz}_0)$
\STATE Take gradient step and update PAE parameters with the loss function $\mathcal{L}_{\text{PAE}} = \|\bw_0 - \widetilde{\bw}_0\|^2$
\ENDFOR
\STATE Use PAE encoder to produce latent parameter representation $\bz_0 = \text{P-Encoder}(\bw_0)$.
\end{algorithmic}
\end{algorithm}

\begin{algorithm}[H]
\caption{Training Algorithm of $\bepsilon_\btheta$ in G-LDM} \label{algo: gldm}

\textbf{Input:} Latent parameter representation $\bz_0 \sim q_{\bz}(\bz_0)$; graph signals $\bX$; graph structure $\bA$; number of diffusion steps $T$; noise schedule $\beta = \{\beta_1, \beta_2, ..., \beta_T\}$; number of training epochs $E_{\text{G-LDM}}$.

\textbf{Output:} The learned denoising network $\bepsilon_\btheta$.

\begin{algorithmic}[1]
\STATE Sample $\bz_0 \sim q_{\bz}(\bz_0)$
\FOR{$i = 1, 2, ..., E_{\text{G-LDM}}$}
\STATE $t \sim \mathrm{Uniform}(1,T), \bepsilon \sim \mathcal{N}(\bzero,\bI)$
\STATE Take gradient step and update $\btheta$:
\begin{equation*}
    \nabla_{\btheta} \sE_{t,\bz_0\sim q_{\bz}(\bz_0),\bepsilon \sim \gN(\bzero,\bI)} \left\| \bepsilon - \bepsilon_\btheta\left(\sqrt{\widetilde{\alpha}_t} \bz_0 + \sqrt{1-\widetilde{\alpha}_t} \bepsilon, t, \bc_{\gG}\right) \right\|^2
\end{equation*}
\ENDFOR

\end{algorithmic}
\end{algorithm}

\begin{algorithm}[h!]
\caption{Inference Algorithm of G-LDM} \label{algo: sample}
\textbf{Input:} Number of diffusion steps $T$; noise schedule $\beta = \{\beta_1, \beta_2, ..., \beta_T\}$; diffusion sampling variance hyperparameter $\sigma = \{\sigma_1, \sigma_2,...,\sigma_T\}$.

\textbf{Output:} Generated latent parameters $\hat{\bz}_0$.
\begin{algorithmic}[1]
\STATE Randomly generate white noises $\hat{\bz}_T \sim \gN (\bzero,\bI)$
\FOR{$t = T, T-1, ..., 1$}
\STATE $\be = 0$ if $t=1$ else $\be \sim \gN (\bzero,\bI)$
\STATE Compute and update
\begin{equation*}
    \hat{\bz}_{t-1} = \frac{1}{\sqrt{1-\beta_t}}\left(\hat{\bz}_t - \frac{\beta_t}{\sqrt{1-\widetilde{\alpha}_t}} \bepsilon_\btheta(\hat{\bz}_t,t,\bc_\gG)\right) + \sigma_t \be
\end{equation*}
\ENDFOR

\end{algorithmic}
\end{algorithm}

\section{Details of PAE and G-LDM Denoising Network Architecture} \label{apx: conv_arch}

\begin{figure}[h!] 
    \centering
    \includegraphics[height = 5cm, width=0.55\linewidth]{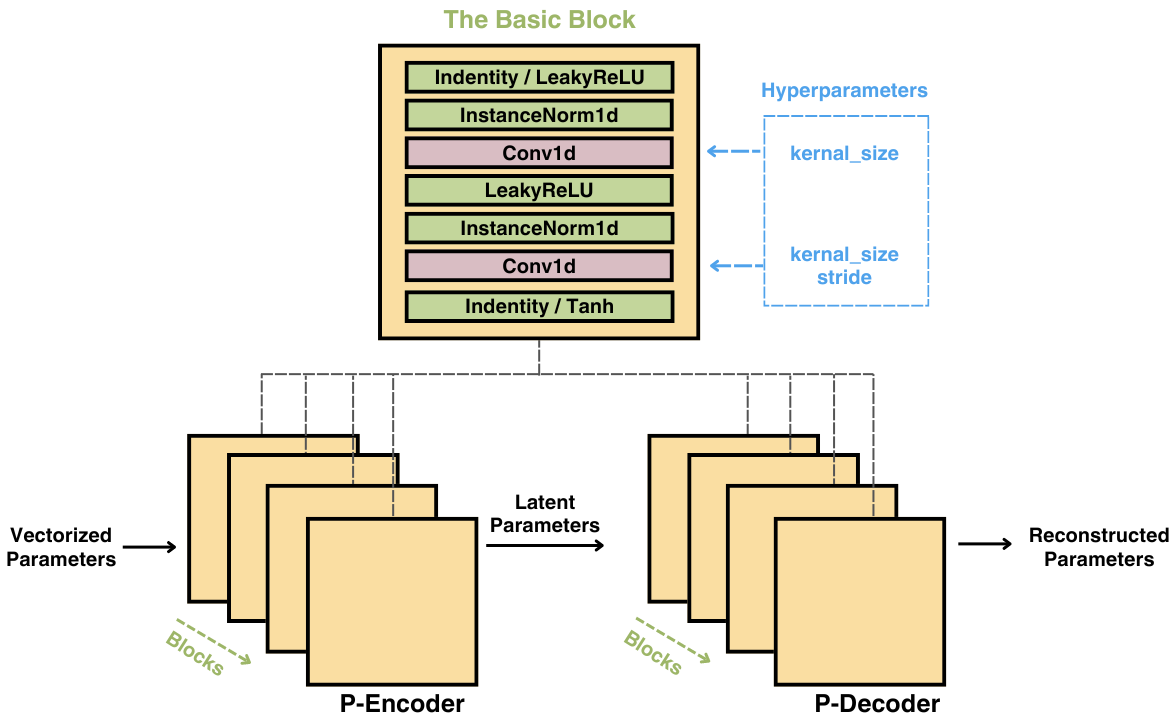}
    \caption{PAE architecture. Hyperparameters required tuning are from 2 \texttt{Conv1d} layers.}
    \label{fig:pae}
\end{figure}

\begin{figure} [h!]
    \centering
    \includegraphics[height = 5.5cm,width=0.55\linewidth]{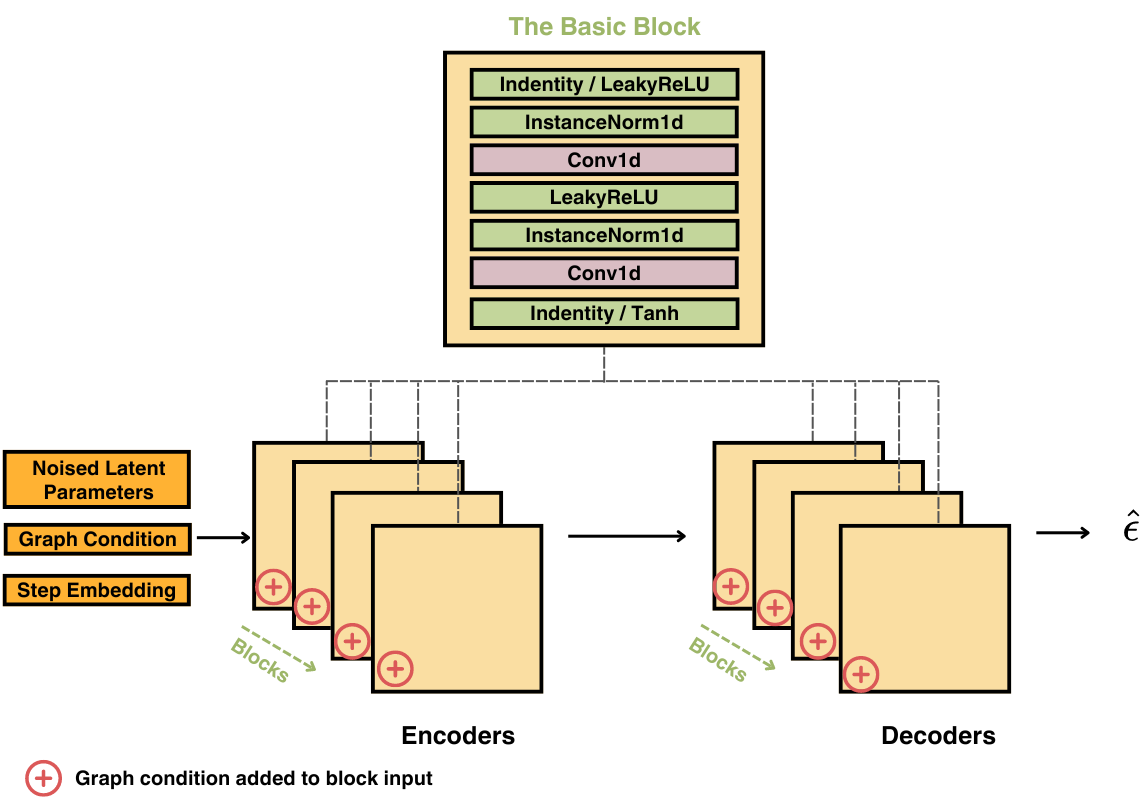}
    \caption{$\bepsilon_\btheta$ architecture. We mark the positions where additional graph condition is involved.}
    \label{fig:g-ldm}
\end{figure}

We provided visualized PAE and G-LDM $\bepsilon_\btheta$ architectures in Figure \ref{fig:pae} and Figure \ref{fig:g-ldm}, respectively. Taking the idea from \cite{wang2024neural}, both architectures have \texttt{Conv1d}-based layers with LeakyReLU activation and instance normalization. Hyperparameters that need to be tuned during training of GNN-Diff are from \texttt{Conv1d} layers in PAE. More information will be provided in Appendix \ref{apx: pae_hyper}. In Figure \ref{fig:g-ldm}, we mark the positions of graph condition $\bc_\mathcal{G}$ being involved. We include graph condition in each encoder and decoder block by adding it to the input of the block directly. For more details of implementation, please refer to the code attached.

\section{Details of Benchmark Datasets} \label{apx: data}

Datasets are retrieved from \texttt{PyTorch Geometric}. Relevant statistics are provided in Table \ref{tab: data}. 

\begin{table}[h!]
\centering
\caption{Statistics of benchmark datasets. ``\#'' denotes ``number of''.}
\renewcommand{\arraystretch}{1.1}
\scalebox{0.9}{
\begin{tabular}{c|ccccc}
\hline
Dataset   & Type         & \# Nodes & \# Edges & \# Features & \# Node Classes \\ \hline
Cora      & Homophilic   & 2708     & 10556    & 1433        & 7               \\ \hline
Citeseer  & Homophilic   & 3327     & 9104     & 3703        & 6               \\ \hline
Actor     & Heterophilic & 7600     & 30019    & 932         & 5               \\ \hline
Chameleon & Heterophilic & 2277     & 62792    & 2325        & 5               \\ \hline
\end{tabular}}
\label{tab: data}
\end{table}

\section{Details of GNN Models and Why We Choose Them}\label{apx:GNN}

\textbf{GCN} We start by recalling the popular Graph Convolution Networks (GCN) \cite{kipf2016variational} with the feature propagation rule: 
\begin{align}
    \text{GCN}:  \mathbf X^{(\ell + 1)} = \sigma \big( \mathbf A \mathbf X^{(\ell)} \mathbf W^{(\ell)}  \big), \label{eq_classic_gcn}
\end{align}
in which we denote $\mathbf A \in \mathbb R^{N\times N}$ as the adjacency matrix of the graph, and $\mathbf W$ as the weight matrix used for channel-mixing. GCN model propagates the graph node signal by spatially aggregating its neighboring information via the graph adjacency matrix, serving as the very first attempt to leverage the deep learning tools for handling the graph-structured data. 

\textbf{GAT} Building upon the scheme provided by GCN, the Graph Attention Networks \cite{velivckovic2017graph} introduces an attention mechanism to enhance the performance of GCN by considering the node feature information. Specifically, we have: 
\begin{align}
    \text{GAT}:  \mathbf {X}^{(\ell+1)} = \sigma (\boldsymbol{\Theta}^{(\ell)} \odot \mathbf A \mathbf X^{(\ell)} \mathbf W^{(\ell)}), \label{GAT} 
\end{align}
where we denote $\boldsymbol{\Theta}_{i,j}$ as the  attention score on edge $i,j$. We note that there are various approaches to generating $\boldsymbol{\Theta}$; in this paper, we only consider the method proposed in \cite{velivckovic2017graph}, with multi-head attention. Furthermore, compared to GCN \eqref{eq_classic_gcn}, which only uses a linear operator ($\mathbf A$) on the graph signals, whereas in GAT, in each layer propagation, the entries of $\boldsymbol{\Theta}$ will depend on the current graph features, thus in general, $\boldsymbol{\Theta}$ tends to be different between layers, suggesting a non-linear feature propagation. 

\textbf{APPNP} Nevertheless, the so-called over-smoothing (OSM) issue is still observed via both models \cite{lee2023towards}. The OSM problem refers to the observation that node features tend to be identical after a certain propagation of GNN layers, suggesting an exponential decay with respect to a non-similarity measure on the node features \cite{rusch2023survey}. In addition, the OSM problem of GNN is closely related to GNN's adaption power on the heterophily graph since relatively higher feature variation is required to let GNN produce distinct label predictions. Many attempts have been made to counter the aforementioned problems, and one of the most popular paradigms is to bring back the variation of the features by adding the so-called source term via GNN propagation, resulting a range of GNN models \cite{han2024from,zhai2024bregman,shi2024revisiting}. We start by reviewing the APPNP model \cite{gasteiger2019predict} which owns the feature propagation as: 
\begin{align}
    \text{APPNP}: \mathbf X^{(\ell+1)} = (1-\alpha)\widehat{\mathbf A}\mathbf X^{(\ell)} + \alpha \mathbf X^{(0)}, \quad \mathbf X^{(\mathrm{out})} = \mathbf X^{(L)} \mathbf W,
\end{align} 
where the feature information is bought back by adding additional source term $\mathbf X(0)$, and $\mathbf W \in \mathbb R^{D_f \times D_c}$ which projects the feature from initial feature dimension $D_f$ to the number of classes $D_c$ after $L$ times propagation via APPNP. 

\textbf{SAGE} Similar to APPNP, the graph sample and aggregate (SAGE) model \cite{hamilton2017inductive} is with the propagation rule as:
\begin{align}
    \text{SAGE}: \mathbf X^{(\ell +1)} = \mathbf X^{(\ell)}\mathbf W_1^{(\ell)} + \mathbf A\mathbf X^{(\ell)}\mathbf W_2^{(\ell)},  
\end{align}
where the source term is generated from the (embedded) feature matrix in addition to the original GCN propagation. 

\textbf{H2GCN} Apart from bringing source terms to handle the OSM and heterophily problem, H2GCN \cite{zhu2020beyond} propagates the node features by aggregating higher-order neighboring information. Specifically, for any node $i$, its initial feature $\mathbf X_i(0)$ is firstly embedded by a learnable weight matrix $\mathbf W^{(0)}$ and then after $\ell$ iterations we have: 
\begin{align}
    \text{H2GCN}: &\mathbf X_{i,h}^{(\ell+1)} = \mathrm{AGGR}\{\mathbf X_j^{(\ell)} : j\in \mathcal N_h(i)\} = \sum_{j \in \mathcal N_h(i)} \mathbf X_i^{(\ell)}d_{i,h}^{1/2}d_{j,h}^{1/2}, \text{and} \\ 
    & \mathbf X^{\mathrm{final}}_{i} = \mathrm{Combine}(\mathbf X_i^{(0)}, \mathbf X_i^{(1)},\cdots \mathbf X_i^{(L)}), 
\end{align}
where we let $h$ be the neighboring order (hops) that H2GCN uses to aggregate, and the representation is formed as a combination of all intermediate representations. Lastly, another embedding of $\mathbf X^{\mathrm{final}}$ is leveraged to make the final prediction of the labels. 

\textbf{ChebNet} Along with the development of these classic spatial GNNs, spectral GNNs, which propagate node features via spectral filtering, have also gained popularity. The initial research spectral GNNs can be found in the so-called ChebNet model \cite{defferrard2016convolutional}, with the propagation rule as: 
\begin{align}
    \text{ChebNet}: \mathbf X^{(\ell +1)} = \sum_{m=0}^K \mathcal T_m(\widetilde{\mathbf L})\mathbf X^{(\ell)} \mathbf W^{(\ell)},
\end{align}
where $\gT_m(\widetilde{\mathbf L})$  is the Chebyshev polynomial of order $m$ evaluated at $\widetilde{\mathbf L} = \frac{2}{\lambda_{\mathrm{max}}}\mathbf L - \mathbf I$, where $\lambda_{\mathrm{max}}$ is the largest eigenvalue of the graph Laplacian $\mathbf L$. We highlight that although many advances \cite{shao2023unifying,shi2023curvature,shi2024design} have been made to enhance the original ChebNet due to the OSM and heterophily problems, ChebNet still serves as the most fundamental basics of deploying spectral GNNs. 

\textbf{GIN} Finally, the last type of GNNs we included in our generative scheme is GNNs that aim to maximize the graph-level expressive power. It has been proved that the standard GCN model (i.e., equation \eqref{eq_classic_gcn})) is with the distinguishing power no higher than the so-called WL test \cite{xu2019how}. To address this problem, the Graph Isomorphism Networks (GIN) is defined as: 
\begin{align}
    \text{GIN}: \mathbf X^{(\ell +1)} = \mathrm{MLP}2 \left((\mathbf A+ (1+ \xi)\cdot \mathbf I)\cdot \mathbf X^{(\ell)}\right),
\end{align}
where $\xi$ is the initial random perturbation on the graph adjacency matrix. 
We highlight that we use the operator $\mathrm{MLP}2$ instead of denoting it via a single weight matrix since to maximize the expressive power of GNN, one layer of $\mathrm{MLP}$ is not enough according to \cite{xu2019how}.

\noindent \textbf{Summary} In this section, we illustrate four types of GNN models: classic spatial GNNs, spectral GNNs, GNNs with source terms, and GNNs with maximized expressive power. Although many other GNN models have been proposed in recent years, these four types of GNNs are typical enough to represent the GNN families, thus sufficient to illustrate the enhancement power of our proposed generative schemes.

\section{Search Spaces of Grid Search and Coarse Search} \label{apx: search space}
\subsection{Coarse search}
Optimizer and training hyperparameters:
\begin{itemize}
    \item Optimizer: [SGD, Adam]
    \item SGD momentum: [0.9]
    \item Learning rate: [1, 0.5, 0.1, 0.05, 0.005]
    \item Weight decay: [5e-3, 5e-4]
    \item Dropout: [0.5] 
\end{itemize}

Model architecture and other hyperparameters:
\begin{itemize}
    \item GCN1: None
    \item GCN2: Hidden size = [16, 64]
    \item APPNP: Teleport probability $\alpha$ = [0.1, 0.5, 0.9]
    \item SAGE: Hidden size = [16, 64]
    \item ChebNet: Hidden size = [16, 64], $K$ = 2
    \item GIN: Hidden size = [16, 64]
    \item GAT: Hidden size = [16, 64], num\_heads = [4, 8]
    \item H2GCN: Hidden size = [16, 64]
\end{itemize}

\subsection{Grid search}
Optimizer and training hyperparameters:
\begin{itemize}
    \item Optimizer: [SGD, Adam]
    \item SGD momentum: [0.9]
    \item Learning rate: [1, 0.5, 0.1, 0.05, 0.01, 0.005]
    \item Weight decay: [1e-3, 5e-3, 1e-4, 5e-4]
    \item Dropout: [0.1, 0.5, 0.9] 
\end{itemize}

Model architecture and other hyperparameters:
\begin{itemize}
    \item GCN1: None
    \item GCN2: Hidden size = [16, 32, 64]
    \item APPNP: Teleport probability $\alpha$ = [0.1, 0.3, 0.5, 0.7, 0.9]
    \item SAGE: Hidden size = [16, 32, 64]
    \item ChebNet: Hidden size = [16, 32, 64], $K$ = [1, 2, 3]
    \item GIN: Hidden size = [16, 32, 64]
    \item GAT: Hidden size = [16, 32, 64], num\_heads = [4, 6, 8]
    \item H2GCN: Hidden size = [16, 32, 64]
\end{itemize}

\subsection{Number of Configurations}

We list the number of configurations of coarse search and grid search for target GNNs in Table \ref{tab: num_config}. It is clear that the search spaces of grid search are much broader than coarse search. The number of configurations in grid search space is around 7 to 30 times of that in coarse search space.

\begin{table}[h!]
\centering
\renewcommand{\arraystretch}{1.1}
\caption{Number of configurations in coarse search and grid search for target GNNs.}
\scalebox{0.9}{
\begin{tabular}{c|cccccccc}
\hline
Method        & GCN1 & GCN2 & APPNP & SAGE & ChebNet & GIN & GAT  & H2GCN \\ \hline
Coarse search & 20   & 40   & 60    & 40   & 40      & 40  & 80   & 40    \\ \hline
Grid search   & 144  & 432  & 720   & 432  & 1269    & 432 & 1269 & 432   \\ \hline
\end{tabular}}
\label{tab: num_config}
\end{table}

\section{Details of GNN-Diff Training} \label{apx: pae_hyper}
\textbf{Training} The total number of training epochs is $E = E_{\text{GAE}} + E_{\text{PAE}} + E_{\text{G-LDM}}$, with $E_{\text{GAE}} = 200$, $E_{\text{PAE}}$ between $15000$ and $60000$ depending on the parameter size, and $E_{\text{G-LDM}} = 45000$. GAE is trained with the same graph data as coarse search. PAE is trained with vectorized parameter samples, while G-LDM learns from their latent representation. Batch size for PAE and G-LDM training is 50. All three modules are trained with the same AdamW optimizer \cite{loshchilov2017decoupled} with learning rate 1e-3 and weight decay 2e-3. 

\textbf{GAE} The concatenation dimension $D_\gG$ is set the same as the latent parameter dimension $D_p$, which is decided by the \texttt{Conv1d} layer in PAE. Dropout is applied before linear operator $\bW_4$ and $\bW_5$ with dropout rate $0.1$.

\textbf{PAE} P-Encoder and P-Decoder are both composed of 4 basic blocks in Figure \ref{fig:pae}. Since the input is latent representation of vectorized parameters, the number of input channel of the first block in P-Encoder is 1. The number of output channels for \texttt{Conv1d} layers in P-Encoder are $6$. Accordingly, the number of input channels of P-Decoder is also $6$. The number of output channels of P-Decoder are [512,512,8,1]. To ensure high quality generation, we observe in experiments that the latent parameter dimension is best between $48$ and $96$. It is worth noting that the latent parameter is actually the concatenation of $6$ output channels of P-Encoder, which means the output dimension for each channel is best between $8$ and $16$. This is done by appropriately setting the stride value. Besides, kernal size needs to be properly tuned for good performance as well. To minimize the effort caused by hyperparameter tuning in GNN-Diff, we provide a fairly simple rule to decide suitable configurations. We set the same value for stride and kernel rate, and decide this value based on the original parameter size of target GNN. For example, we set both stride and kernel size as $5$ for a parameter vector containing 5k to 10k parameters. When multiple values can reduce the parameter dimension to $8$ to $16$, usually choosing the one that reduces to lower dimension (i.e., $8$ to $10$) leads to better generation results. In Table \ref{tab: pae_hyper}, we provide the configurations we used for different target GNNs on various datasets.

\begin{table}[h!]
\centering
\renewcommand{\arraystretch}{1.1}
\caption{Hyperparameter configurations of PAE.}
\scalebox{0.9}{
\begin{tabular}{c|ccc|ccc}
\hline
\rowcolor[HTML]{EFEFEF} 
Dataset & \multicolumn{3}{c|}{\cellcolor[HTML]{EFEFEF}Cora}  & \multicolumn{3}{c}{\cellcolor[HTML]{EFEFEF}Citeseer}  \\ \hline
Model   & Parameter size      & Stride     & Kernal size     & Parameter size       & Stride      & Kernal size      \\ \hline
GCN1    & 10038               & 5          & 5               & 22224                & 7           & 7                \\
GCN2    & 23063               & 6          & 6               & 59366                & 9           & 9                \\
APPNP   & 10038               & 5          & 5               & 22224                & 7           & 7                \\
SAGE    & 46103               & 8          & 8               & 118710               & 9           & 9                \\
ChebNet & 46103               & 8          & 8               & 118710               & 10          & 10               \\
GIN     & 96447               & 9          & 9               & 241648               & 12          & 12               \\
GAT     & 93780               & 9          & 9               & 238792               & 12          & 12               \\
H2GCN   & 23264               & 7          & 7               & 59536                & 9           & 9                \\ \hline
\rowcolor[HTML]{EFEFEF} 
Dataset & \multicolumn{3}{c|}{\cellcolor[HTML]{EFEFEF}Actor} & \multicolumn{3}{c}{\cellcolor[HTML]{EFEFEF}Chameleon} \\ \hline
Model   & Parameter size      & Stride     & Kernal size     & Parameter size       & Stride      & Kernal size      \\ \hline
GCN1    & 4665                & 5          & 5               & 11630                & 6           & 6                \\
GCN2    & 60037               & 9          & 9               & 37301                & 7           & 7                \\
APPNP   & 4665                & 5          & 5               & 11630                & 6           & 6                \\
SAGE    & 120005              & 10         & 10              & 74581                & 9           & 9                \\
ChebNet & 30005               & 7          & 7               & 298309               & 13          & 13               \\
GIN     & 15315               & 6          & 6               & 37603                & 8           & 8                \\
GAT     & 124920              & 11         & 11              & 150332               & 11          & 11               \\
H2GCN   & 15152               & 6          & 6               & 37440                & 8           & 8                \\ \hline
\end{tabular}}
\label{tab: pae_hyper}
\end{table}

\textbf{G-LDM} We set the number of diffusion steps $T = 1000$. We choose the linear beta schedule with $\beta_1 = 1\text{e-}4$ and $\beta_T = 2\text{e-}2$. The denoising network $\bepsilon_\btheta$ adopts an encoder-decoder structure with 8 basic blocks in Figure \ref{fig:g-ldm}. The number of input channels and output channels, stride, and kernal rate are the same as in \cite{wang2024neural}. Since PAE has conducted dimensionality reduction on parameters, and the input of G-LDM, latent parameter representation, has sufficiently low dimension regardless of the target GNN architecture and the dataset, we can use the same hyperparameter configuration in $\bepsilon_\btheta$ to ensure generation quality.

\end{document}